%% file: draft.tex
\documentclass{article}

\usepackage{arxiv}
\usepackage{graphicx}
 
\usepackage[utf8]{inputenc} 
\usepackage[T1]{fontenc}    
\usepackage{hyperref}       
\usepackage{url}            
\usepackage{booktabs}       
\usepackage{amsfonts}       
\usepackage{nicefrac}       
\usepackage{microtype}      
\usepackage{lipsum}
\usepackage{amsmath}
\usepackage{placeins}
\usepackage{wrapfig}
\usepackage{lineno}

\newcommand{\xx}{{\textbf{x}}}
\newcommand{\nn}{{\textbf{n}}}
\newcommand{\un}[1]{\ensuremath{\ \mathrm{#1}}}

\DeclareMathOperator*{\argmin}{argmin}

\title{Estimation of fibre architecture and scar in myocardial tissue using electrograms: an in-silico study}

\author{
\begin{flushleft}
\textbf{
Konstantinos ~Ntagiantas\textsuperscript{1*},
Eduardo ~Pignatelli\textsuperscript{1},
Nicholas S.~Peters\textsuperscript{2},
Chris D.~Cantwell\textsuperscript{3},
Rasheda A.~Chowdhury\textsuperscript{2\$},
Anil A.~Bharath\textsuperscript{1\$}
}

\bigskip
1 Department of Bioengineering, Imperial College London, London SW7 2AZ, United Kingdom
\\
2 National Heart and Lung Institute, Imperial College London, London W12 0NN, United Kingdom
\\
3 Department of Aeronautics, Imperial College London, London SW7 2AZ, United Kingdom
\\
\$ Authors contributed equally.

\bigskip
* Corresponding Author\\
E-mail: konstantinos.ntagiantas19@imperial.ac.uk\\
Address: Royal School of Mines, Department of Bioengineering, Imperial College London, London SW7 2AZ, United Kingdom\\
\end{flushleft}
}

\begin{document}
\maketitle

\input{0_abstract}

\input{1_introduction}

\input{2_methods}

\input{4_results}


\input{5_discussion}

\input{6_conclusion}

\bibliographystyle{unsrt}  
\bibliography{references}

\end{document}

%% file: 0_abstract.tex
\begin{abstract}

Atrial Fibrillation (AF) is characterized by disorganised electrical activity in the atria and is known to be sustained by the presence of regions of fibrosis (scars) or functional cellular remodeling, both of which may lead to areas of slow conduction. Estimating the effective conductivity of the myocardium and identifying regions of abnormal propagation is therefore crucial for the effective treatment of AF.
We hypothesise that the spatial distribution of tissue conductivity can be directly inferred from an array of concurrently acquired contact electrograms (EGMs). We generate a dataset of simulated cardiac AP propagation using randomised scar distributions and a phenomenological cardiac model and calculate contact EGMs at various positions on the field. EGMs are enriched with noise extracted from biological data acquired in the lab. A deep neural network, based on a modified U-net architecture, is trained to estimate the location of the scar and quantify conductivity of the tissue with a Jaccard index of $91$\%. We adapt a wavelet-based surrogate testing analysis to confirm that the inferred conductivity distribution is an accurate representation of the ground truth input to the model. We find that the root mean square error (RMSE) between the ground truth and our predictions is significantly smaller ($p_{val}<0.01$) than the RMSE between the ground truth and surrogate samples. \\
\\
\end{abstract}

%% file: 1_introduction.tex
\section{Introduction}
\label{sec:intro}

The normal propagation of electrical signals through the myocardium leads to coordinated contraction of the heart muscle. The cardiac action potential (AP) reflects the movement of ions between the interior of the myocytes and the extracellular space. When the transmembrane potential of these cells increases above the threshold of activation, an AP takes place \cite{plonsey_barr}. The propagation of APs through the heart is the result of the collective expression of ion channels and gap junctions (connexin proteins) whose roles are, respectively, to signal and initiate the cellular AP and to propagate this AP to neighboring cells. The elongated shape of myocytes that form the cardiac fibres, and the polar positioning of gap junctions in the direction of the fibre orientation leads to anisotropic conduction. Pathological loss of gap junctions or fibre disarray, for example in AF, can lead to changes in the anisotropic ratio. Physiological and pathological heterogeneities and changes in the expression of these proteins in the myocardial cells result in a reduction in effective conductivity and the substrate being non-homogeneous throughout the myocardium \cite{rasheda_2018}. 

In addition to channel abnormalities, areas devoid of myocytes (scars) exhibit lower conductivity and abnormal propagation of the cardiac AP at the macro-scale, potentially generating re-entrant electrical waves which can initiate atrial fibrillation (AF) \cite{fenton_chaos}. Destroying the partially conductive tissue (ablation) may provide an effective way to eliminate slow conduction pathways and reduce the likelihood of reentrant circuits forming \cite{Parameswaran2021}.

EGMs can be measured clinically during cardiac ablation procedures, to investigate arrhythmias and steer treatment \cite{star_af_original}. In contrast to an electrocardiogram (ECG) which is recorded by electrodes places on the skin, the electrogram (EGM) is recorded by electrodes in direct contact with the myocardium. It measures the superposition of electric fields generated by the aforementioned movement of ions in the local field of view of the electrode; it is consequently affected by the heterogeneity and anisotropy of the myocardial substrate and the expression of gap junctions. Furthermore, it provides information about localised changes in conduction. 
The morphology of EGMs has previously been qualitatively binarised (complex fractionated atrial EGMs vs. simple EGMs) or quantified, using techniques such as dominant frequency analysis \cite{Li2016}, by clinicians to identify possible ablation targets with ambiguous results \cite{cfaes_2, cfaes}. This lack of methodological efficacy has prevented widespread implementation of such techniques. Consequently, the current success rate of AF treatment through catheter ablation remains considerably low. Furthermore, the use of modern high-density mapping catheters means that even with a good understanding of EGMs and how their morphology correlates to the underlying structure, the observation and analysis of the multiple signals from these catheters would still be challenging to clinicians, with respect to time and reproducibility, without any algorithmic processing. A better understanding of the underlying substrate, and robust methods for identification of abnormalities, may lead to more successful treatments to combat AF.

In this study, we aim to demonstrate the capabilities of  modern deep learning techniques to infer the structural properties of the tissue, namely the fibre orientation and the tissue conductivity, underlying a set of electrodes used to acquire unipolar EGMs. This validation of technique in a controlled in silico model is a vital step before implementation on clinical data.
Deep learning is being increasingly explored, in place of more traditional techniques, for inferring the solution to inverse problems that quickly become intractable because of ill-posedness or complexity \cite{Lu2021, Li2020, Jin2017, takamoto2022pdebench}. With classical machine learning algorithms like Principal Component Analysis, high-dimensional signals are represented by low-dimensional feature vectors.
By inferring general laws from experimental data, deep learning allows for the enrichment and/or re-evaluation of many of the existing heuristics that are considered informative to solve the inverse problem. Specifically, in the problem we are addressing, there is no consensus about which features of unipolar EGMs are informative of substrate properties and how these signal characteristics can be used to estimate properties of the myocardium. Deep neural networks work by \emph{compressing} the raw, high-dimensional observations into a lower-dimensional space, allowing to treat previously intractable problems. Finally, there is rising evidence that the operator represented by the deep network projects the raw signal not only to a smaller space, but also one that makes the problem linear \cite{Lusch2018}.
Therefore, we propose to fit to a user-generated dataset, a deep neural network, whose input is the entire raw EGM signal. This allows the transition from relying to manually engineered features \cite{Brook2020}, to identifying important EGM parts and morphology during the training process \cite{cantwell_review}. Furthermore, the network is trained to learn the spatial relationship between neighbouring electrodes, through the use of convolutions, to infer the local anisotropy of the substrate. In cardiology, deep learning techniques have been successfully used with electrocardiogram (ECG) data for automatic analysis and diagnosis \cite{Ebrahimi2020}. Hybrid datasets of simulated and clinical intracardiac EGMs have been explored in their ability to classify patient tissue as fibrotic or non-fibrotic \cite{Sanchez2021}. However to the best of our knowledge, there has been no published attempt to quantitatively predict the anisotropic and heterogeneous diffusivity tensor at every point of an unseen field/tissue, in simulated or biological data. In the present work, we predict unseen, simulated diffusivity tensor fields from the respective simulated EGMs.

%% file: 2_methods.tex
\section{Methods}
\label{sec:methods}

\begin{figure}[ht]
    \centering
    \includegraphics[width=1\textwidth]{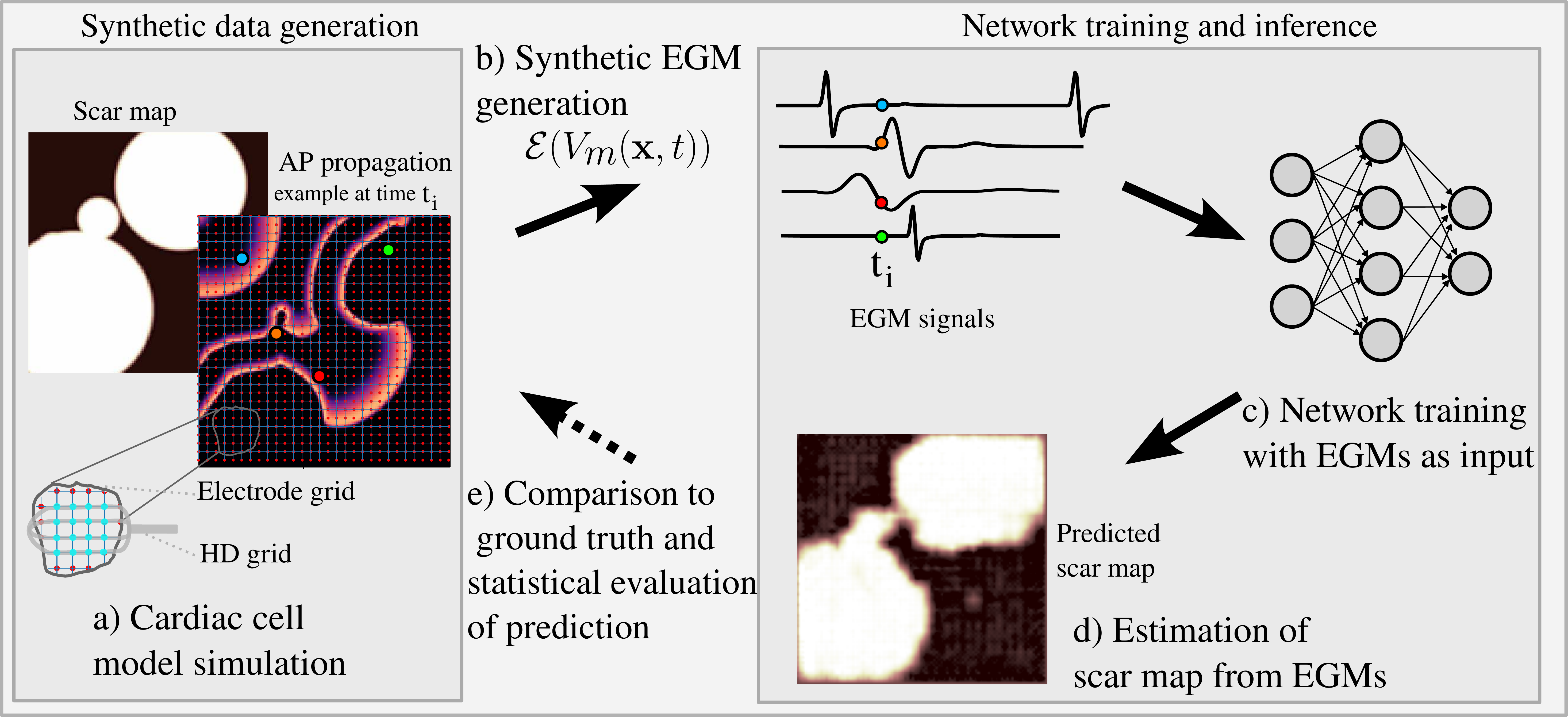}
    \caption[Method]{Schematic of the pipeline. \textbf{a - b}: First we generate the data by solving an AP propagation model in a specific domain characterized by a diffusivity tensor. From the AP simulation, we calculate the EGMs on an \textit{electrode grid}, which has a density similar to the HD grid as shown in the magnification. The calculation is described in detail in Section \ref{subsec:ccm}. Virtual electrodes record EGMs simultaneously. \textbf{c - e}: We use the generated EGMs in the training of a neural network that estimates the scar map of the domain from which the EGMs have been recorded, and evaluate the predictions using statistical methods.}
    \label{fig:method_schematic}
\end{figure}

The outline of the steps we follow are described below and are illustrated diagrammatically in Figure~\ref{fig:method_schematic}. First, we generate a \textit{scar maps} set of synthetic myocardial tissue, consisting of fields with a variety of diffusivity tensors (scar maps). A description of how we generate the scar maps is detailed in Section \ref{subsec:scar_maps}. Next, on each of these generated fields, we simulate AP propagation by solving a cardiac EP model, described below in Section \ref{subsec:ccm}, and compute synthetic EGMs by placing virtual electrodes on top of the field, in a grid arrangement consistent with a clinical \textit{HD grid} catheter. The final dataset is a set of input-output pairs, where one input is a set of EGM time series, and the output its corresponding scar-map. The mathematical formulation of this inverse problem is:

\begin{align}
\label{eq:math_dl_egm}
    \mathcal{F}(D(\xx);V_m(\xx, t)) &= 0,   \quad \xx \in \Omega,  \\
    \mathcal{B}(V_m(\xx)) &= 0,  \quad \xx \in \partial\Omega ,
\end{align}
where, $\Omega$ is the finite computational domain with boundary $\partial\Omega$,  $\mathcal{F}(\cdot)$ is the monodomain model, $V_m$ is the transmembrane potential, $D(\xx)$ is the true diffusivity tensor field, and $\mathcal{B}(\cdot)$ are the imposed boundary conditions. The goal is to then train a network, 
\begin{equation}
\label{eq:math_dl_map}
    h : \mathcal{E}(V_m(\xx,t)) \mapsto \tilde{D}(\xx), \
\end{equation}
where $\mathcal{E}(\cdot)$ is the EGM calculation shown in Equation \ref{eq:met_egm}, and $h$ is a deep neural network with parameter set $\theta$ that maps the EGMs to an estimate of the diffusivity tensor $\tilde{D}(\xx)$.
We train the network using the EGM time series, $\mathcal{E}(V_m(\xx;t)) \in \mathbb{R}^{c\times c \times T}$, to find $h^*(\mathcal{E}, \theta^*)$ with the optimal set of parameters $\theta^*$ that predicts the diffusion tensor fields $\tilde{D}(\xx) \in \mathbb{R}^{n\times n \times 3}$, where the third dimension corresponds to the three components of the symmetric diffusion tensor. Finally, $n \times n $ is the dimension of the discrete field and $c \times c$ is the number of electrodes, each of which provides an EGM sequence, and $T$ is the duration of the input. EGMs are sampled from multiple electrodes concurrently.

\subsection{Cardiac EP model}
\label{subsec:ccm}

We aim to validate our hypothesis in a simulated setting. For this purpose, we use the simplified three-variable Fenton-Karma-Cherry model \cite{Fenton2002}, to produce simulations in a synthetic $12 \times 12$ \un{cm} square region, $\Omega$. The monodomain partial differential equation that governs the action potential propagation, denoted by the function $\mathcal{F}$ in Equation~\ref{eq:fk_simple}, is given by:
\begin{equation}
\label{eq:fk_simple}
    \frac{\partial V_m(\xx, t)}{\partial t} = \nabla \cdot ( D(\xx)\nabla V_m ) + \frac{I_{ion}+I_{stim}}{C_m}
\end{equation}
where the diffusion term $\nabla \cdot ( D(\xx)\nabla V_m )$ propagates the AP through the myocardium, and the reaction term $I_{ion}$ is responsible for generating the APs by modelling the opening and closing of the ion gates. The diffusivity tensor is related to the conductivity as $D(\xx) = \frac{1}{\beta C_m}G(\xx)$, where $\beta$ is the cellular surface-to-volume ratio and $C_m = 1 \un{\mu F/cm^2} $ \cite{silvio,conduction} represents the membrane capacitance. $I_{stim}$ corresponds to the injected stimuli current to initiate and maintain AP propagation. In the Fenton-Karma-Cherry model, $I_{ion}$ is modelled with three currents and three state variables, one of which is the transmembrane voltage $V_m$. Although mathematically simple, this model can accurately reproduce ventricular AP propagation and so provides a favourable starting point from a methodological perspective.

It is given by the equations:

\begin{align*}
    I_{ion} &=  - ({J_{fi}(u , v) + J_{so}(u) + J_{si}(u, w)}),\\
    \frac{\partial v}{\partial t} &= \mathcal{H}(u_c - u)(1-v)/\tau_v^-(u) - \mathcal{H}(u-u_c)v/\tau_v^+,\\
    \frac{\partial w}{\partial t} &= \mathcal{H}(u_c - u)(1-w)/\tau_w^- - \mathcal{H}(u-u_c)w/\tau_w^+,\\
    J_{fi}(u, v) &= - \mathcal{H}(u-u_c)(1-u)(u-u_c)(v/\tau_d),\\
    J_{so}(u)    &= \mathcal{H}(u_c-u)(u/\tau_0) + \mathcal{H}(u-u_c)(1/\tau_r),\\
    J_{si}(u, w) &= -(1+ \tanh(k(u-u_c^{si})))w/(2\tau_{si}),
\end{align*}
where $J_{fi}$ is a fast inward current responsible for the depolarisation of the membrane, $J_{so}$ is a slow outward current responsible for the repolarisation of the membrane, and $J_{si}$ is a slow inward current that opposes $J_{so}$ in the recovery phase. We use $\mathcal{H}$ to denote the Heaviside step function. $u$ is the dimensionless form of $V_m$, while $v$ and $w$ are auxiliary variables that represent the biophysical state of each unit of myocardium alongside $u$. To revert from non-dimensional forms, the relation is given by $V_m = u(V_{fi}-V_0)+V_0$, where $V_0$ is the membrane resting potential and $V_{fi}$ the reversal potential of $I_{fi}$. Similarly, $J_{i} = I_i/(C_m(V_i-V_0))$ for $i=fi, so, si$.

We apply Neumann boundary conditions $\mathcal{B}$ at the boundaries of the square area:

\begin{align}
     \frac{dV_m}{d\nn}\Bigr|_{\partial\Omega} = 0 \qquad
\end{align}
where $\nn$ is the direction normal to the boundary. 

\begin{figure}[tb]
    \centering
    \includegraphics[width=0.8\textwidth]{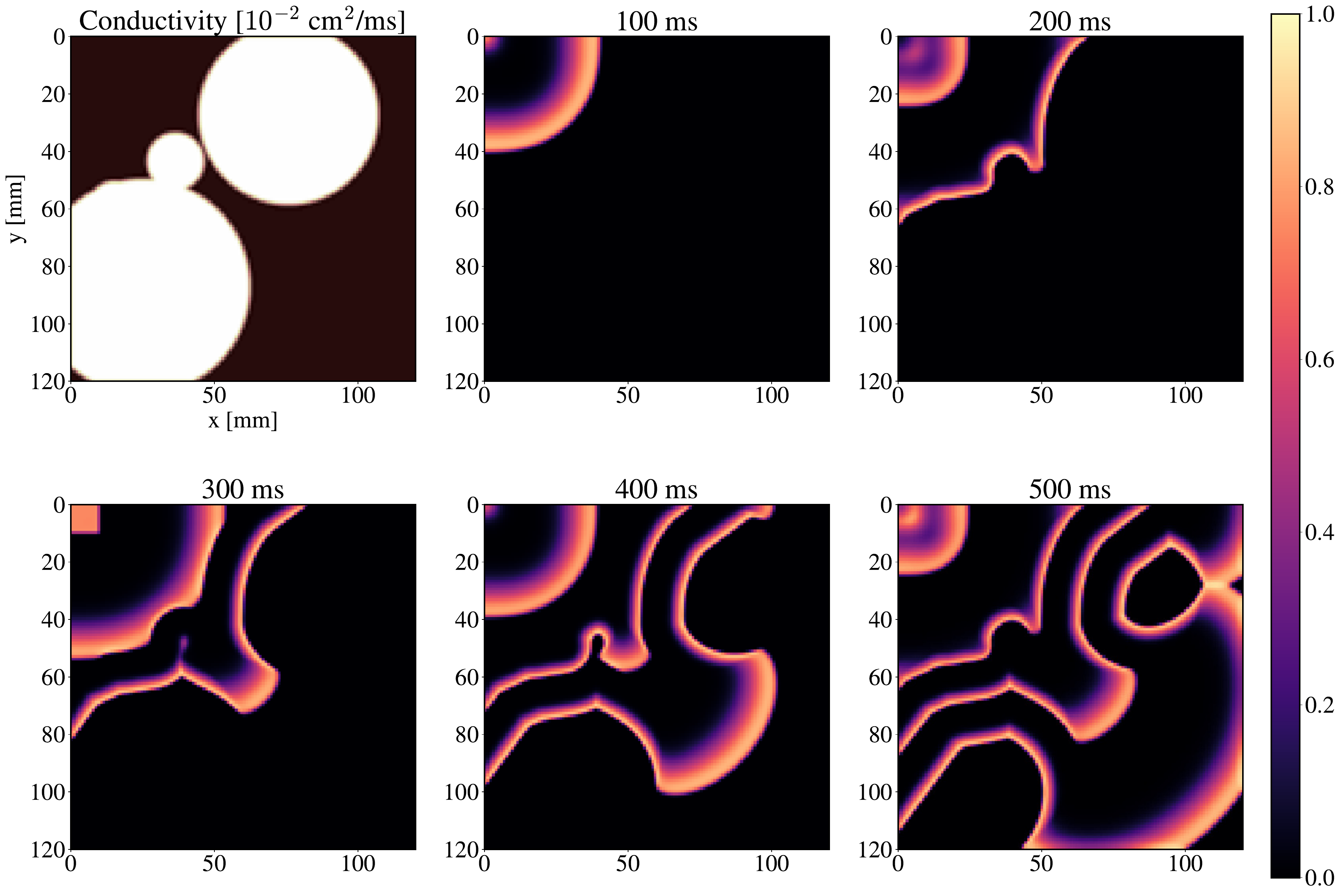}
    \caption[Simulation Example]{Diffusivity field and corresponding AP propagation in different instances of a simulation. The modelled scar is shown in white, and is $90\%$ less conducting than healthy tissue ($D = 10^{-3} \un{cm^2/ms}$ for healthy tissue and $D = 10^{-4} \un{cm^2/ms}$ for modelled scar respectively. The effect of the low conductivity region can be seen in all of the frames shown, slowing down the wavefront. For this particular example, we stimulate the tissue with a point stimulus at the top left corner of the domain, resulting in a circular wavefront. All values are normalized from zero to one.}
    \label{fig:simulation_example}
\end{figure}

In the isotropic setting where the tensor field is a scalar multiple $d$ of the identity, we use $d = 10^{-3} \un{cm^2/ms}$ for the healthy tissue \cite{clayton_review}, while for the scar regions of low conductivity $d = 10^{-4} \un{cm^2/ms}$. An example of a diffusion field and the corresponding propagation of the AP through the domain can be seen in Figure \ref{fig:simulation_example}. Our numerical solver is implemented using the software JAX for GPU-accelerated computations \cite{jax}. A second-order accurate finite differences scheme is used for computing spatial gradients, and a forward Euler scheme for temporal derivatives, with $dx = dy = 0.01 \un{cm}$ and $dt = 0.01 \un{ms}$ respectively. The resulting simulations have a spatial resolution of $1200 \times 1200$ grid points, representing a tissue of $12 \un{cm} \times 12 \un{cm}$.

\subsection{Electrogram calculation}
\begin{figure}[tb]
    \centering
    \includegraphics[width=1\textwidth]{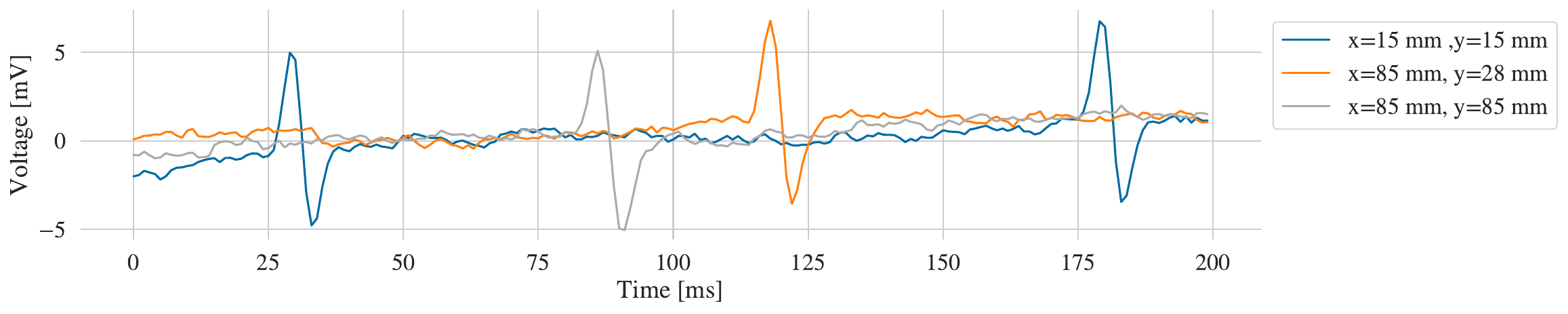}
    \caption[EGM Example]{Contact electrogram signals acquired from different positions in the field shown in Figure \ref{fig:simulation_example}. The second channel from position $x = 85 \un{mm},\quad y=28 \un{mm}$ has a significantly lower output, since the electrode is positioned within the scar region. Noise from a biological noise has been added to the signals.}
    \label{fig:egm_example}
\end{figure}

In the clinical setting, EGM signals are recorded from the endocardium using mapping catheters. For the purposes of this study, the EGM signals are calculated from the solutions to the monodomain equation, at specific positions on the field. The extracellular potential $\phi_e$ is the sum of currents in the field, weighted by the inverse square distance from the electrode \cite{plonsey_barr}, given by the surface integral:

\begin{equation}
\label{eq:met_egm}
     \mathcal{E}(V_m(\xx, t)) = \phi_e(\xx, t) = \int_\Omega\frac{\nabla V_m(\xx^\prime, t) \cdot (\xx - \xx^\prime)}{4\pi\sigma_{e}||\xx-\xx^\prime||^3} \,d\Omega
\end{equation}
where $\nabla V_m(\xx^\prime, t)$ is the spatial gradient of the transmembrane voltage in $\Omega$, at time $t$, $\xx$ is the location of the probe, and $\sigma_e$ is the conductivity of the domain, with a typical value $\sigma_e = 20 \un{mS/cm}$ \cite{stinstra,conduction}. The virtual probes are assumed to have infinitesimal spatial extent (i.e. the virtual electrode radius is zero).
For the calculation of the simulated EGM, a second-order accurate finite differences scheme is used for the gradient of the voltage field, and the rectangle rule for numerically calculating the integral over $\Omega$. The virtual electrode is placed at a vertical distance of $z=1 \un{mm}$ above the field to reduce computational artefacts and avoid a singularity. An example of EGM signals acquired at different positions for the simulation shown in Figure \ref{fig:simulation_example}, can be seen in Figure \ref{fig:egm_example}. 

We calculate the EGM signal at a uniform rectangular grid of points, at a sampling rate of $1 \un{ms}$. This \emph{electrode grid} is coarser than the finite differences grid used to solve the monodomain model, and the choice of an inter-electrode spacing of $4 \un{mm}$ is inspired by modern grid catheters used in the clinical setting for contact EGM acquisition \cite{abbott_1, abbott_2}. Each simulation results in 841 EGMs signals in total.

\subsection{Biological Noise Model for Simulated EGMs}
\label{subsec:noise}

The cardiac EP model and the EGM calculation described above are fully deterministic, and the resulting signals are clean without any noise sources. Biological EGM recordings from the lab environment or the clinic are particularly noisy signals, due to contraction, contact and farfield artefacts, even after standard preprocessing that involves removing measurement noise like the $50\un{Hz}$ power line interference. This makes it necessary to create a noise model for the simulated EGMs. Training and testing with appropriately \emph{noisy} EGMs demonstrates that neural networks may realisticaly be translated to a biological/ clinical setting even with non-optimal data.

The noise model we create is a deterministic autoregressive process that is generated from biological EGM signals acquired in the pre-clinical laboratory from \emph{ex-vivo} Langendorff-perfused porcine hearts \cite{Brook2020}, at a recording rate of $1\un{kHz}$. The EGMs are recorded with the clinical Abbott HD Grid catheter  \cite{abbott_1, abbott_2}, which is compromised of 36 electrodes, arranged in a $4 \times 4$ grid with the same density between adjacent electrodes as we use in the simulated electrode grid. A drawing of the catheter and how it resembles our grid can be seen in the lower left corner of Figure \ref{fig:method_schematic}. 

\begin{figure}[tb]
    \centering
    \includegraphics[width=1\textwidth]{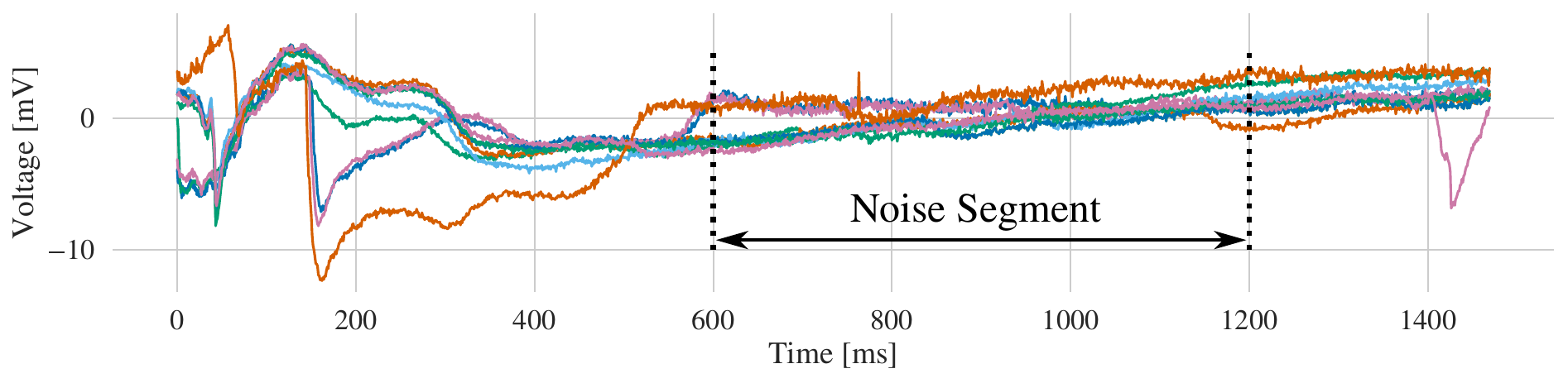}
    \caption[Biological EGMs and noise]{Examples of biological unipolar EGMs that were used to create the noise model. The data comes from ex-vivo Langendorff pig hearts, paced at a cycle length of $1000\un{ms}$. The parts that were considered to create the noise model are contained within the two vertical dashed lines.}
    \label{fig:real_egms}
\end{figure}

\begin{wrapfigure}{L}{0.5\textwidth}
    \centering
    \includegraphics[width=0.5\textwidth]{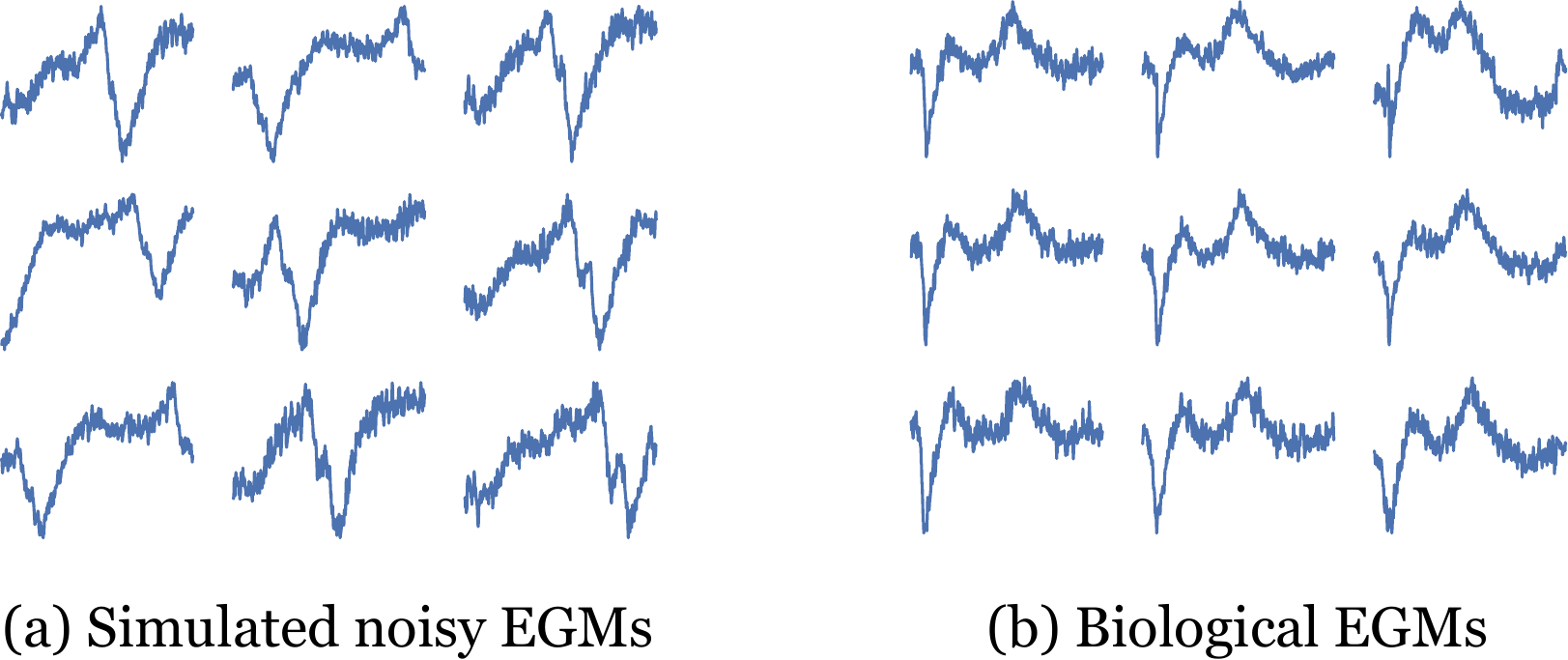}
    \caption[Simulated and Biological EGMs]{A side-by-side example of simulated and biological EGMs. The $3\times 3$ grid is part of the electrode grid in both cases. }
    \label{fig:biological_egms_vs_sim}
\end{wrapfigure}

The noise model is constructed following the steps in \cite{Sanchez2021}. We extract noise segments from $10$ unipolar EGM signals acquired in the pre-clinical laboratory. As noise, we consider the signal after stimuli and subsequent activations, in paced beats, with a pacing frequency of $0.67\un{Hz}$ ($1500\un{ms}$ cycle length). Prior to noise extraction, we filter the $50\un{Hz}$ power line interference that is encountered in the laboratory setting. An example of the noise segments, along with the original unipolar EGMs from which they come, is shown in Figure \ref{fig:real_egms}.

As described in \cite{Sanchez2021}, each segment is fitted with an autoregressive model, and the model coefficients are averaged across segments. Although this is a deterministic model $n(t)$, by randomly sampling $t_{start}$ and $t_{end}$, we generate pseudorandom noise $\mathbf{n} = [n(t_{start}), n(t_{start+1}),\ldots,n(t_{end})]$ that we add to our simulated EGMs. Finally, we add white noise $\epsilon \sim \mathcal{N}(0,0.3)$ after normalizing the EGMs. Examples of this process are shown in \ref{fig:biological_egms_vs_sim}, where the EGMs shown in each frame come from adjacent electrodes. The simulated EGMs are shown after noise has been added, via the process described in the main text. The biological EGMs have the $50\un{Hz}$ noise removed and white noise is added.

\subsection{Fully Convolutional Networks}
\label{subsec:met_network}

\begin{wrapfigure}[23]{L}{0.5\textwidth}
  \begin{center}
    \includegraphics[width=0.5\textwidth]{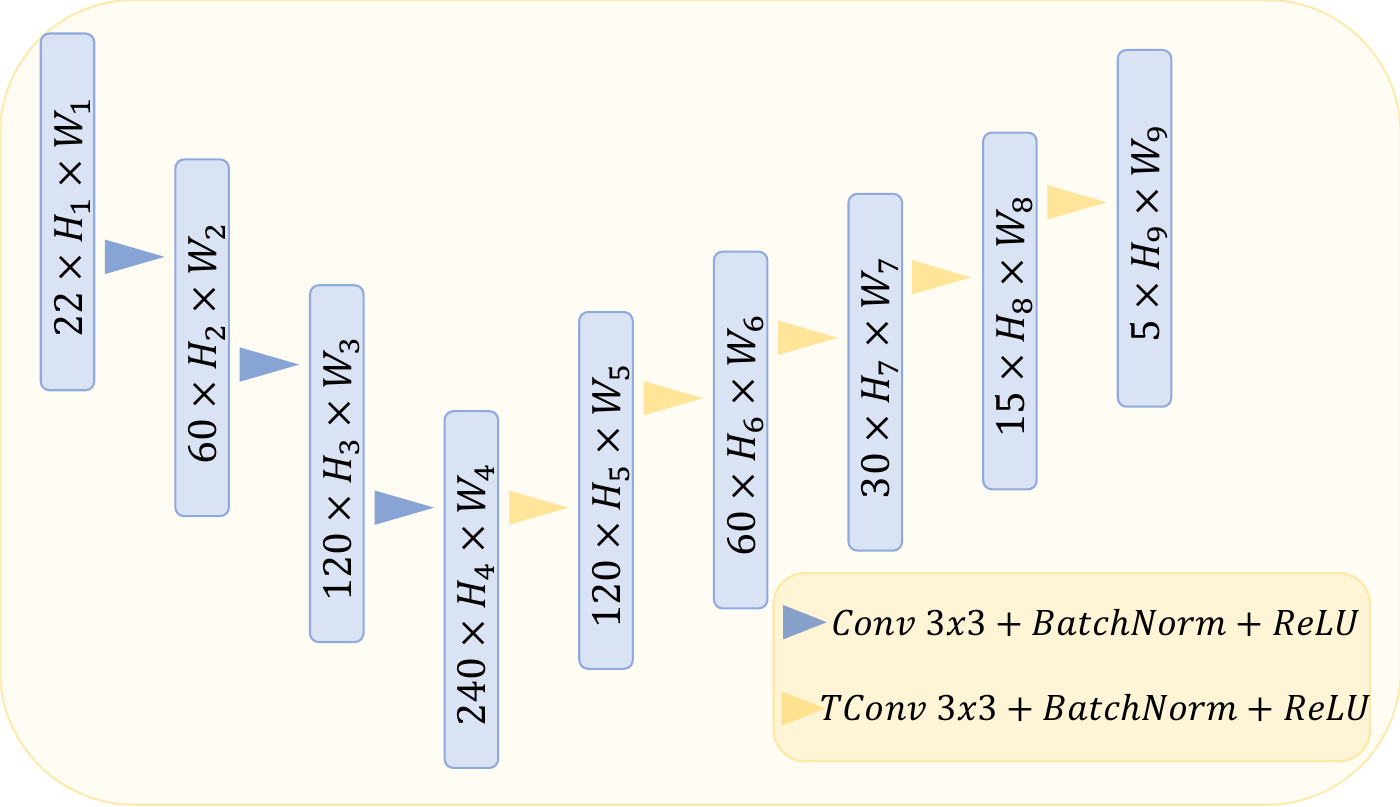}
  \end{center}
  \caption[Architecture]{Architecture of the Encoder-Decoder network that was used. In this configuration, each electrogram signal contains 20 time points. The extra 2 channels of the input are the normalized $x$ and $y$ coordinates of the probes.}
  \label{fig:5_model_architecture}
\end{wrapfigure}

We use a feed-forward fully convolutional neural network. A convolutional layer convolves multiple, parameterised filters (kernels) with the input field, a process that extracts local spatial information from the input data. Since convolution operations are invariant to the size of the inputs, the method is applicable to cardiac tissue of arbitrary sizes. Convolutional networks have been very successful in tasks such as image and instance segmentation for scene parsing or medical imaging analysis \cite{unet}, due to their ability to represent and encode information from spatial fields. We use an Encoder-Decoder architecture, similar to a U-Net \cite{unet}, without the skip connections. The encoder projects the input into subsequent lower-dimensional feature spaces that extract useful spatial and temporal information. These are then parsed through the decoder for upscaling through transposed convolutions and upsampling. The architecture is shown in Figure~\ref{fig:5_model_architecture}.

We use the $H_1\times W_1$ grid of electrogram signals as the input to the model, together with the normalised coordinate of each electrode. The three-dimensional input has height $H_1$ and width $W_1$ corresponding to the size of the electrode grid. Each of the $N$ time points from these signals is a separate channel. 
Since we are dealing with spatiotemporal data, the absolute $x$ and $y$ coordinates of both the EGM probes and the predicted diffusion tensor values are important information to capture in the model. We use a \emph{CoordConv} \cite{liu2018intriguing} layer instead of the traditional convolutional layer in the first and last layers of the model to support this. The \emph{CoordConv} variation is implemented by adding two extra channels to the input and output of the network, that contain the $x$ and $y$ coordinates of the electrode positions (in the input) and the diffusion tensor field units (in the output) respectively.
The first, second and third layers of the encoder extract information from the signal to higher-order feature spaces of depth 60, 120 and 240 respectively, while the depth of the decoder layers is 120, 60, 30, 15 and finally the output has a depth of 5 channels; the first three channels represent the fields $D_{xx}$, $D_{yy}$ and $D_{xy}$ of the diffusion tensor, and the last two channels contain the $x$ and $y$ coordinates of the electrode probe corresponding to each grid unit (\emph{CoordConv} variation). The decoder also upsamples the information to a resolution of $96$ by $96$. Every convolutional layer is followed by a batch normalization layer and a ReLU activation layer. The EGM signals are normalised, the biological EGM noise is added, and Gaussian white noise with a standard deviation of 0.05, $\hat{\phi}_e = \phi_e + \mathcal{N}(0,0.05)$, is added to improve the robustness of the model, where $\phi_e$ is the z-score normalized EGM signal and $\mathcal{N}(0,0.05)$ represents the Gaussian noise. Noise is similarly also added to the \emph{CoordConv} layers.
We train the network for 100 epochs, and use the variation of Adam \cite{kingma2014adam} with weight decay \cite{Loshchilov2017}, with a learning rate of $0.001$ and weight decay of $0.01$. We use the root mean square error (RMSE) as the loss function.
We formalise the optimisation problem as follows:
\begin{equation}
\label{eq:nn_objective}
    h^{*}(\mathcal{E}, \theta) = \argmin_{\theta} \, \mathop{\mathbb{E}}_{(\mathcal{E}, D(\xx)) \sim \mathcal{D}}\left[ \sqrt{\lVert h(\mathcal{E}, \theta) - D(\xx) \lVert^{2}_{2}}\right]
\end{equation}
where $h^{*}(\mathcal{E}, \theta) = h(\mathcal{E}, \theta^{*})$ denotes the network after 100 iterations of Adam, and $\mathcal{D}$ is the dataset as described above.
To evaluate the accuracy of scar estimation, the Jaccard index is used, $J(s, \hat{s}) = \frac{|s\cap \hat{s}|}{|s\cup \hat{s}|}$, where $s, \hat{s}$ is the true scar map and its prediction respectively, expressing the percentage of scar tissue that is identified correctly by the network. During training, we use 10-fold cross validation to assess the performance of the model \cite{hastie01statisticallearning}.

\subsection{Scar and Fibre Maps}
\label{subsec:scar_maps}

\begin{figure}[ht]
    \centering
    \includegraphics[width=0.8\textwidth]{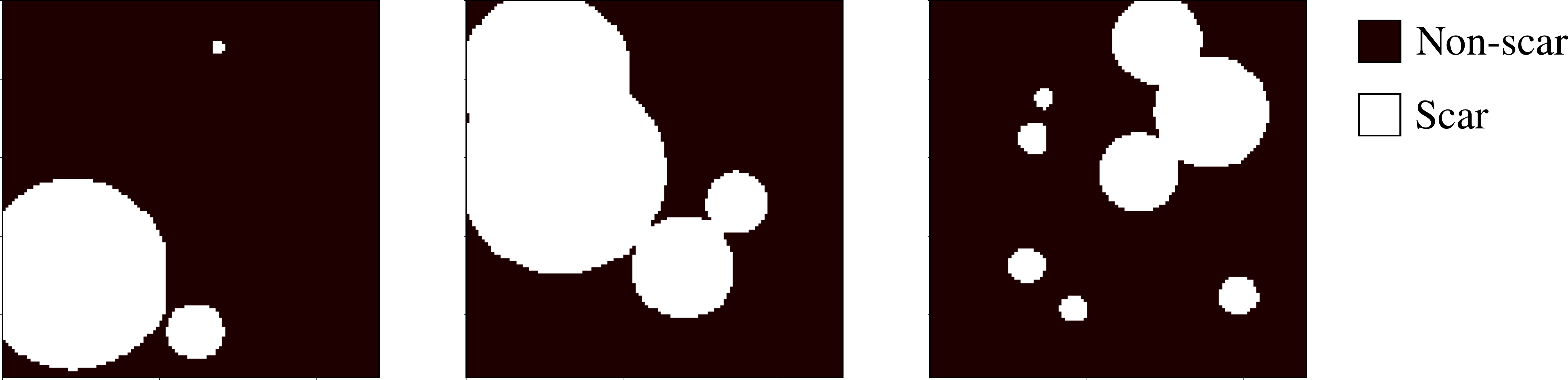}
    \caption[Scar Maps]{Examples of heterogeneous isotropic conductivity fields with compact regions of fibrosis. White regions represent scar and have a lower conductivity value compared to the black regions, which represent healthy tissue.}
    \label{fig:3_scarMaps}
\end{figure}

\begin{figure}[ht]
    \centering
    \includegraphics[width=0.8\textwidth]{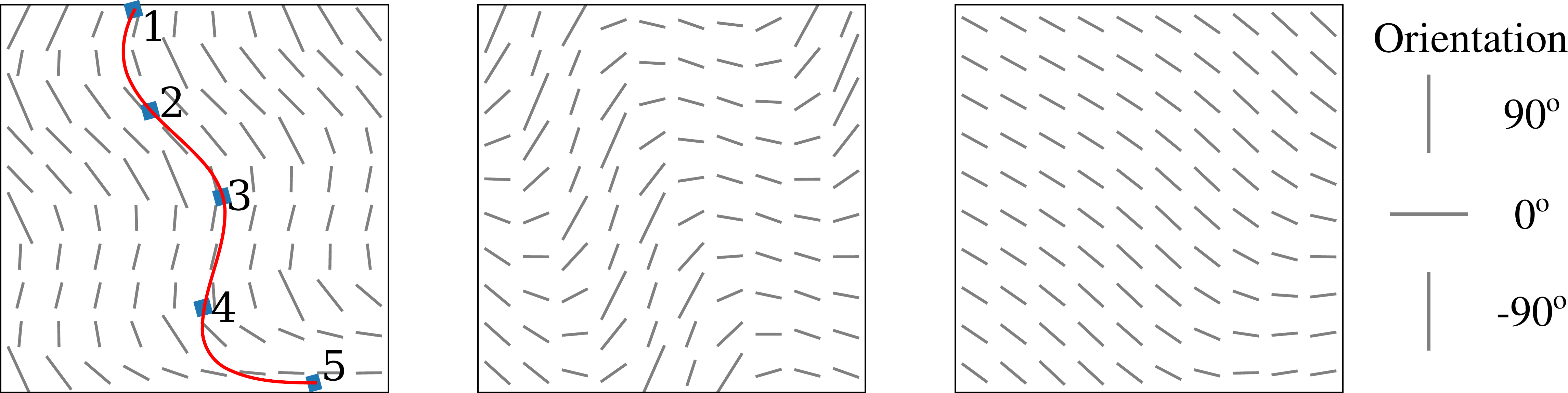}
    \caption[Scar Maps]{Examples of homogeneous anisotropic fibre orientation fields. The lines depict local fibre orientation. On the first example, we show the five control points and resulting spline which is used to generate the scar map. }
    \label{fig:3b_scarMaps}
\end{figure}

Each simulation of the generated dataset is characterized by a different diffusivity tensor field, representing a distribution of scar and fibre orientation. Isotropic heterogeneous, anisotropic homogeneous and anisotropic heterogeneous virtual substrates are considered. In the first two cases, our architecture is evaluated on the estimation of scar location and fibre orientation independently; Figure~\ref{fig:3_scarMaps} and Figure~\ref{fig:3b_scarMaps} show examples of generated maps for these two cases. In the third case, both the scar location and fibre orientation are estimated in combination.
For the isotropic scar maps, the low-conductivity regions are representative of compact fibrosis patterns \cite{Nguyen2014}. Such anatomical barriers can be critical for the onset and perpetuation of AF, as they can be the cause of single reentrant circuits \cite{Allessie1977}. 
Other types of cardiac fibrosis patterns, including the more common interstitial and patchy types, result from collagen bundles separating the excitable myocytes. The degree of fibrosis depends on the ratio of collagen bundles to myocytes. For such cases, the local average of the conductivity would need to be represented in the diffusivity map, but this configuration is beyond the scope of the current work. 

Homogeneous anisotropic maps model the fibre orientation of cardiomyocytes and are used to test the ability of the neural network to estimate it from sparse EGM recordings. Anisotropy is introduced by the differentiation of longitudinal versus transversal conduction in cardiac myocytes; $d_l$ and $d_t$ respectively. To produce the anisotropic maps, first the longitudinal conduction $d_l(\xx)$ is defined across the field, together with the ratio $\lambda(\xx) = \frac{d_l(\xx)}{d_t(\xx)}$ and the fibre angle $\alpha(\xx)$ in degrees. The angle $\alpha(\xx)$ is generated through a rule-based approach \cite{Krueger2011}. First, a path between two points on the field is constructed randomly, defined by a spline passing through five control points ${(x_i, y_i) \quad i=1,...,5}$, where $x_i$ are equidistantly spaced, with $x_1, x_5$ being the two initial points, and $y_i \sim \mathcal{N}(0, 0.09)$. The fibre orientation of the whole field is then calculated by extending the path perpendicularly to the line passing through the two initial points. The selection of five control points for the spline produces fields that are complex enough to robustly asses the performance of the network. Examples of generated anisotropic fields can be seen in the bottom row of Figure \ref{fig:3_scarMaps}. In the homogeneous tissue $\lambda(\xx) = 4$ throughout the field, and $d_l(\xx) = 10^{-3} \un{cm^2/ms}$. In the heterogeneous cases, $d_l(\xx)$ is equivalent to a heterogeneous isotropic scar map.

The diffusion tensor of each simulation is represented by three matrices $D_{xx}$, $D_{yy}$ and $D_{xy}$, with dimensions equal to the dimensions of the field, where each $i, j$-entry contains the corresponding $d_{xx}$, $d_{yy}$ and $d_{xy} = d_{yx}$  values of the diffusion tensor at that location. For isotropic fields, $D_{xx} = D_{yy}$ and $D_{xy} = 0$. These three matrices are the output of the deep neural network.

\subsection{Surrogate Testing Analysis}
\label{subsec:met_surs}

Although the loss metric RMSE can provide some evidence that the model generalizes well, it is not in itself sufficient to evaluate the predictive performance of the network. In order to statistically validate that the predictions made by the network are significant, they are compared to other feasible solutions \cite{Peiffer2013}. We use surrogates to do this \cite{Venema2006}. A simple approach when using surrogate testing methods is to produce surrogates by randomly permuting the elements of the vectors or matrices. However, this method does not preserve the spatial autocorrelation of the scar maps, which in both isotropic and anisotropic cases is high and must be taken into consideration; a random permutation results in non-feasible noisy solutions, leading to an overestimation of the significance of our predictions \cite{Rowland2015}. To create surrogate samples that preserve autocorrelation, we use the dual-tree complex wavelet transform (DT-CWT) which has been successfully used in similar tests \cite{Deblauwe2012}. 

We produce proxy fields from the predictions as feasible outputs of the network, as mentioned in section~\ref{subsec:met_surs}. The proxies represent surrogate data that permit a form of permutation test; each proxy field represents a potential draw of possible values of $D_{xx}$ values over the locations of space corresponding to the output map. Like a permutation test, each of these draws represents the data under the null hypothesis, which is that the model produces estimates of $D_{xx}$ that are unrelated to its true spatial distribution. However, unlike a permutation test, we do not merely scramble the values, but instead select only permutations that have a similar spatial structure to the real target spatial distribution (ground truth map).

%% file: 4_results.tex
\section{Results}
\label{sec:results}

\subsection{Dataset Generation}
\label{subsec:res_data}

The generated dataset consists of 330 simulations, each of which is defined by a different diffusion tensor field: 107 heterogeneous isotropic; 186 homogeneous anisotropic; and 36 heterogeneous anisotropic fields. Examples of the fields used in the simulations are shown in Figure~\ref{fig:3_scarMaps} and Figure~\ref{fig:3b_scarMaps}. Heterogeneous anisotropic fields are obtained as superpositions of heterogeneous isotropic and homogeneous anisotropic fields (scar and variable fibre orientation).

Each simulation has a duration of $1 \un{s}$, paced from a $10 \times 10 \un{mm}$ rectangular region the top left corner of the domain with a pulse-wave signal of period $150 \un{ms}$. The pacing region can be seen in the bottom-left frame of Fig~\ref{fig:simulation_example}. Electrograms are recorded at a frequency of $1 \un{kHz}$ and so the interval between time points is $\Delta t=1 \un{ms}$. Therefore, the available data for every simulation consists of $841$ unipolar electrogram recordings arranged spatially in a $29 \times 29$ \emph{electrode} grid, as shown in Fig~\ref{fig:method_schematic}, with each recording including $1000$ time points. This data is structured in a three-dimensional $1000 \times 29 \times 29$ array.

\begin{figure}[ht]
    \centering
    \includegraphics[width=0.9\textwidth]{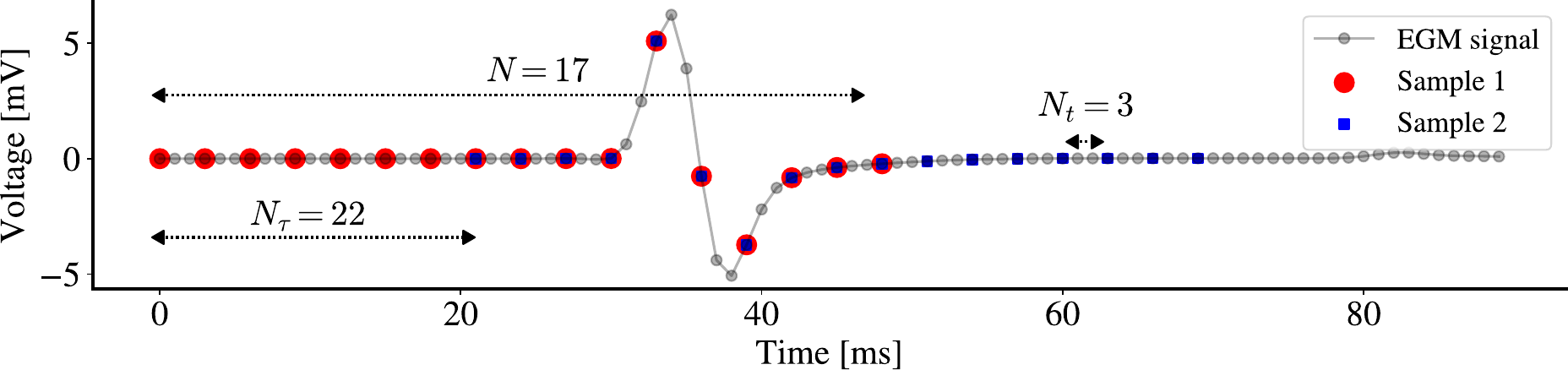}
    \caption[EGM sample selection parameters]{Parameters $N$, $N_t$, and $N_{\tau}$, defining how we sample from the EGM time series.}
    \label{fig:n_nt_ntau}
\end{figure}

\begin{figure}[ht]
    \centering
    \includegraphics[width=0.9\textwidth]{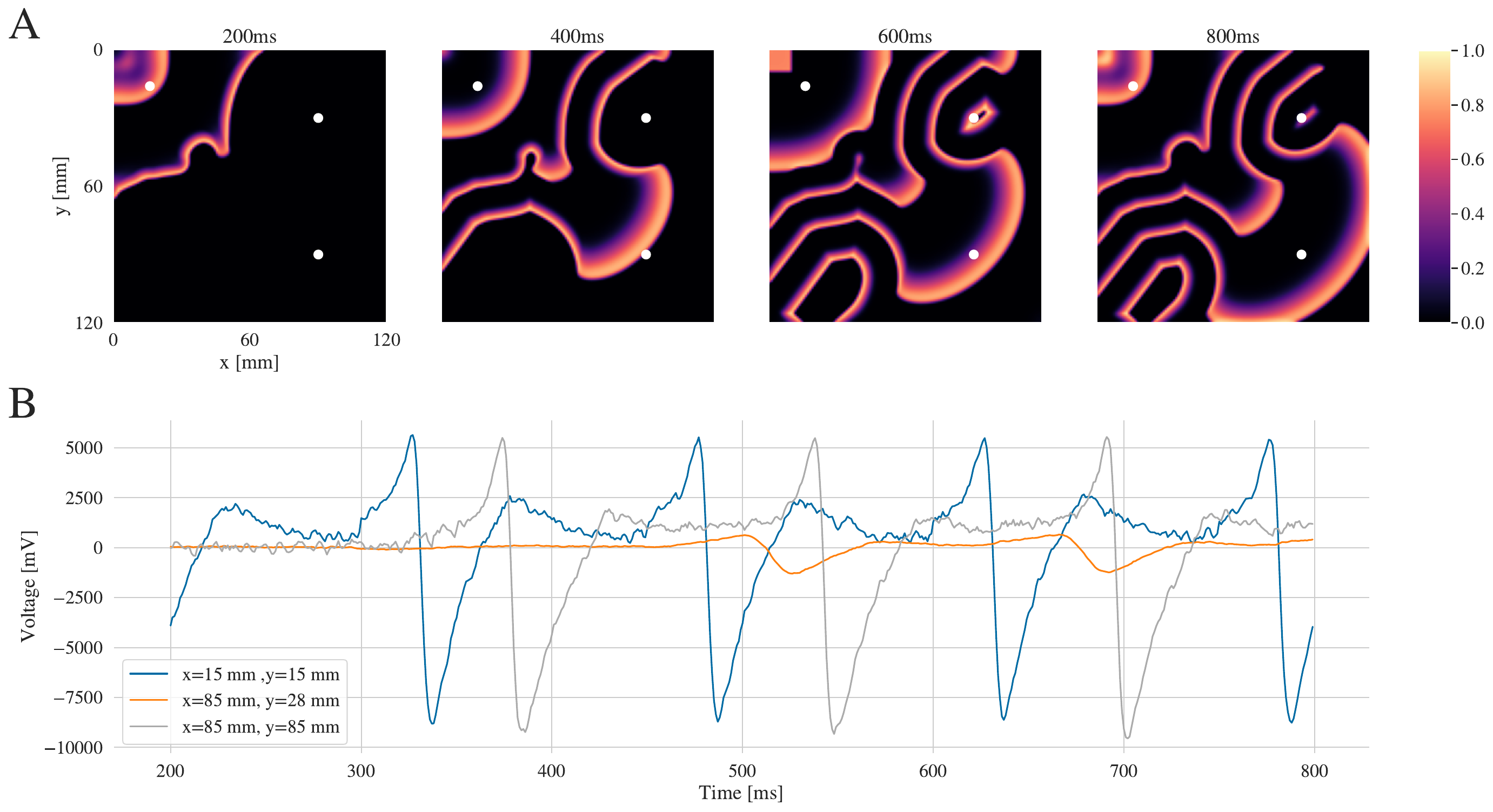}
    \caption[Sample Input]{Sample from the training dataset. \textbf{A}: Sequence of AP propagation sampled from one simulation. The point stimulus is introduced in the top left corner of the field as seen in the first frame. \textbf{B}: Examples of electrogram signals acquired from the simulation shown above. The electrograms are sampled every 1 \un{ms} and for a total of 600 \un{ms} for every sample.}
    \label{fig:4_sample_input}
\end{figure}

Input samples for the network are then obtained by extracting subsets of these data. Three parameters define how these subsets are selected: $N$, the number of time points in each sample; $N_t$, the number of electrogram time points between the time points of the sample; and $N_{\tau}$, the number of time points between the starting points of subsequent samples from the same simulation. A choice of $N_t=1$ will retain each consecutive time-point of the electrograms, while $N_t=2$ will retain every alternate sample from the signal, for example. An example, with $N=17$, $N_t=3$, and $N_{\tau}=22$ can be seen in Fig~\ref{fig:n_nt_ntau}.
Suppose the electrogram signal at electrode $(i,j)$ of simulation $k$, is denoted by
$\phi_e^k(\mathbf{x}_{ij}, t)$, then the $m$-th sample of simulation $k$, $S_m^k$ is defined as

\begin{align*}
S_m^k = \left\{
\phi_e^k\left(\mathbf{x}_{ij}, (mN_{\tau} + nN_t)\Delta t\right) \,|\, n \in [1..N],\, i,j\in[1..29]
\right\}
\end{align*}
where $[ a .. b ]$ denotes the range of integers between $a$ and $b$ inclusive.
Three EGMs from one such sample can be seen in Fig \ref{fig:4_sample_input}, where $N_t=1$, $N = 600$.

We can therefore obtain $\nu = \left \lfloor{\frac{L - (N-1)N_t}{N_{\tau}}+1}\right \rfloor$ EGM input samples, where $L=1000$ is the total number of time points in the simulated electrograms. The model produces one estimation of the diffusion field for every sample, so the final prediction is calculated as the average estimate across all samples from the same simulation.

\subsection{Network Performance}
\label{subsec:res_performance}

\begin{figure}[ht]
    \centering
    \includegraphics[width=1\textwidth]{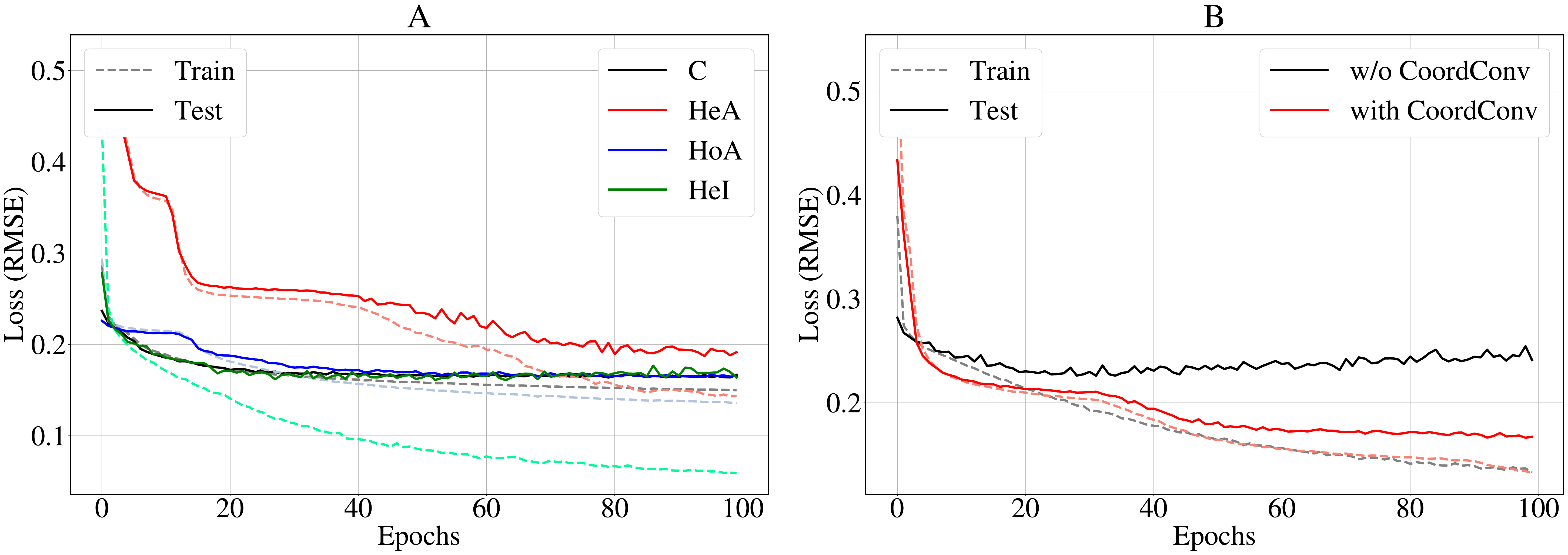}
    \caption[Loss]{Training and validation loss curves for the model. (Left) The different colors correspond to the performance of the model when we consider specific subsets of the \textit{scar map} dataset for the training and testing: C = Combined cases of all fields, HeA = Heterogeneous Anisotropic fields, HoA = Homogeneous Anisotropic fields, HeI = Heterogeneous Isotropic fields.} 
    \label{fig:6_loss}
\end{figure}

The performance of the network in four different cases is evaluated, depending on the type of diffusion field included in the training and testing datasets: we have four different modes where we train and test on heterogeneous isotropic (HeI); homogeneous anisotropic (HoA); heterogeneous anisotropic (HeA); and the combined case where all simulations are used (C). In each of these cases, we do a grid-search optimization for the sampling parameters $N$, $N_t$, $N_{\tau}$ mentioned in the previous section \ref{subsec:res_data}. Based on the average testing set error in the final predicted diffusion tensor fields across all folds, the best performing combination in all the cases is $N = 10, N_t=5, N_{\tau} = 25$. Given the parameter set and the dimensions of the field, the wavefront created from a point stimulus originating on the border of the field takes $200 \un{ms}$ to travel to the opposite side. This implies that a sample spanning $50 \un{ms}$ contains information about approximately a quarter of the field, so the parameter choice makes sense from a physical perspective. The training and loss curves for this best-performing parameter set are presented in Figure \ref{fig:6_loss}A, showing a small generalization error across all cases. As previously mentioned in section \ref{subsec:met_network}, the \emph{CoordConv} layers are implemented to explicitly include spatial information about the position of the electrodes. To validate our selection, the error in the predictions of the network when \emph{CoordConv} is not implemented is increased by $55 \%$, and the respective loss curves are shown in Fig \ref{fig:6_loss}B.

\begin{figure}[ht]
    \centering
    \includegraphics[width=0.9\textwidth]{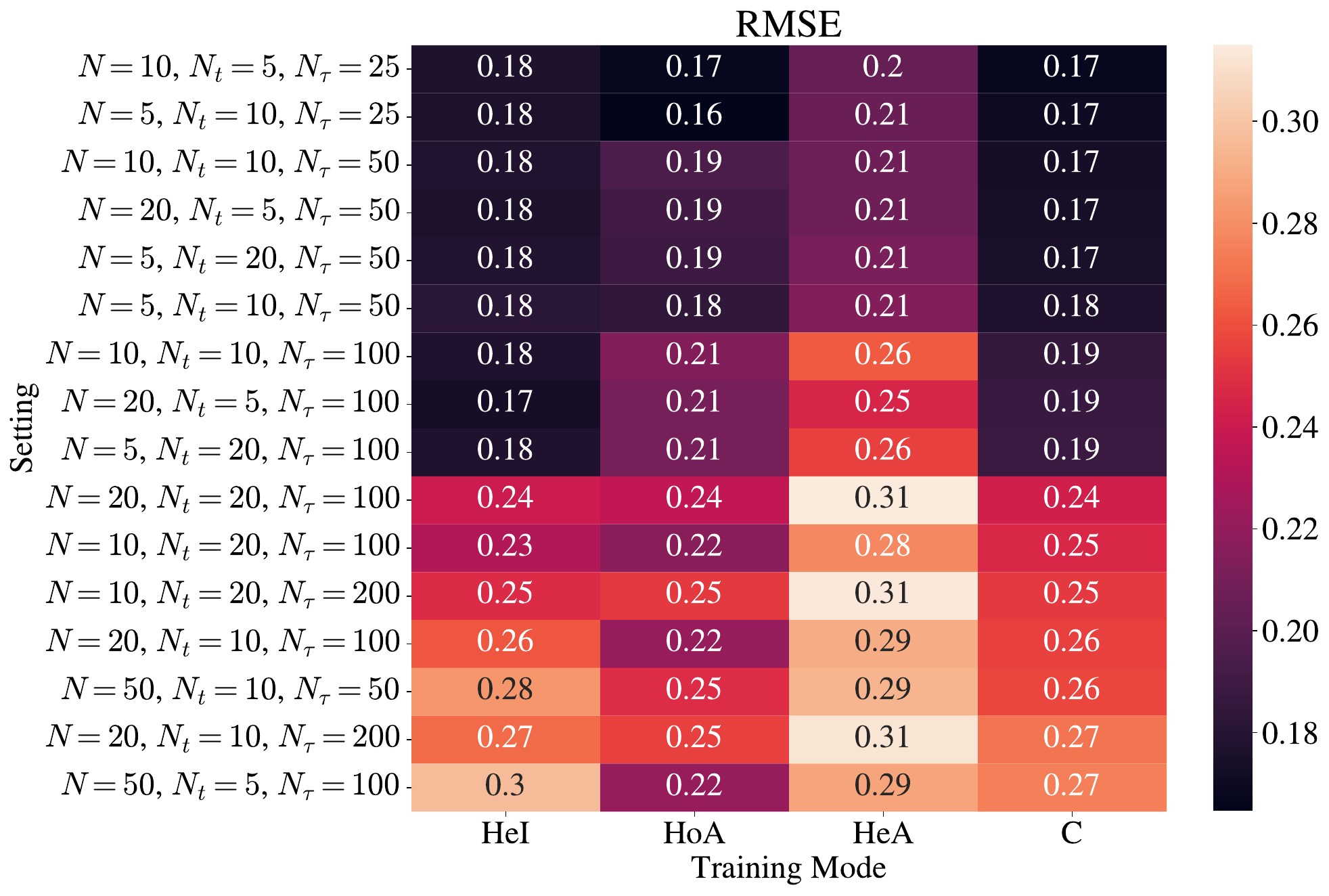}
    \caption[Hyperparams]{Average RMSE across all folds in the testing set. Modes refer to the training and testing \emph{scar map} dataset used: C = Combined cases of all fields, HeA = Heterogeneous Anisotropic fields, HoA = Homogeneous Anisotropic fields, HeI = Heterogeneous Isotropic fields.}
    \label{fig:hyperparams}
\end{figure}

\begin{figure}[ht]
    \centering
    \includegraphics[width=1\textwidth]{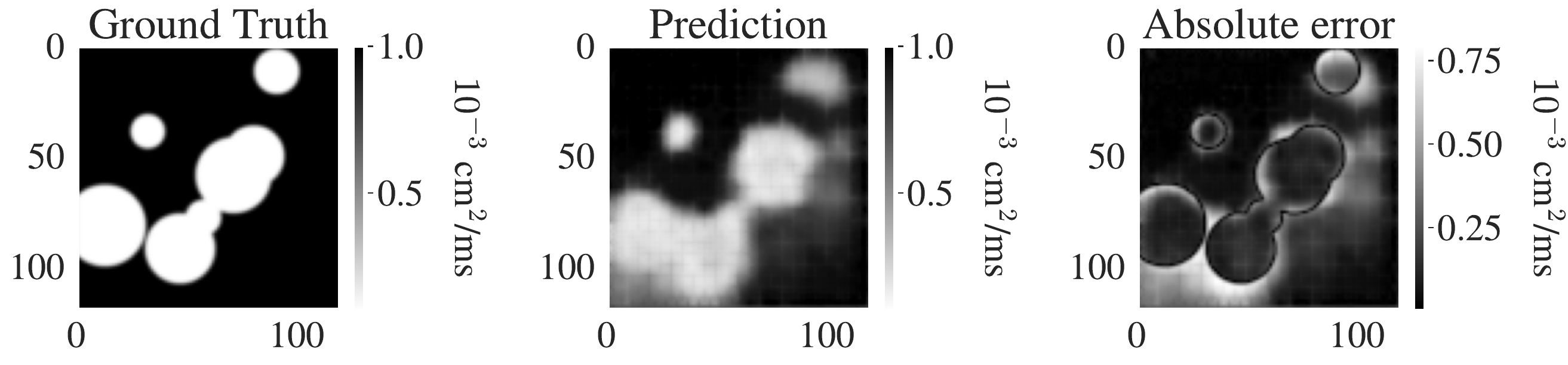}
    \caption[Isotropic]{Predicted isotropic fields with fibrotic regions. The ground truth is shown for comparison. The absolute error is larger near the boundaries of the scar.}
    \label{fig:7_isotropic}
\end{figure}

\begin{figure}[ht]
    \centering
    \includegraphics[width=1\textwidth]{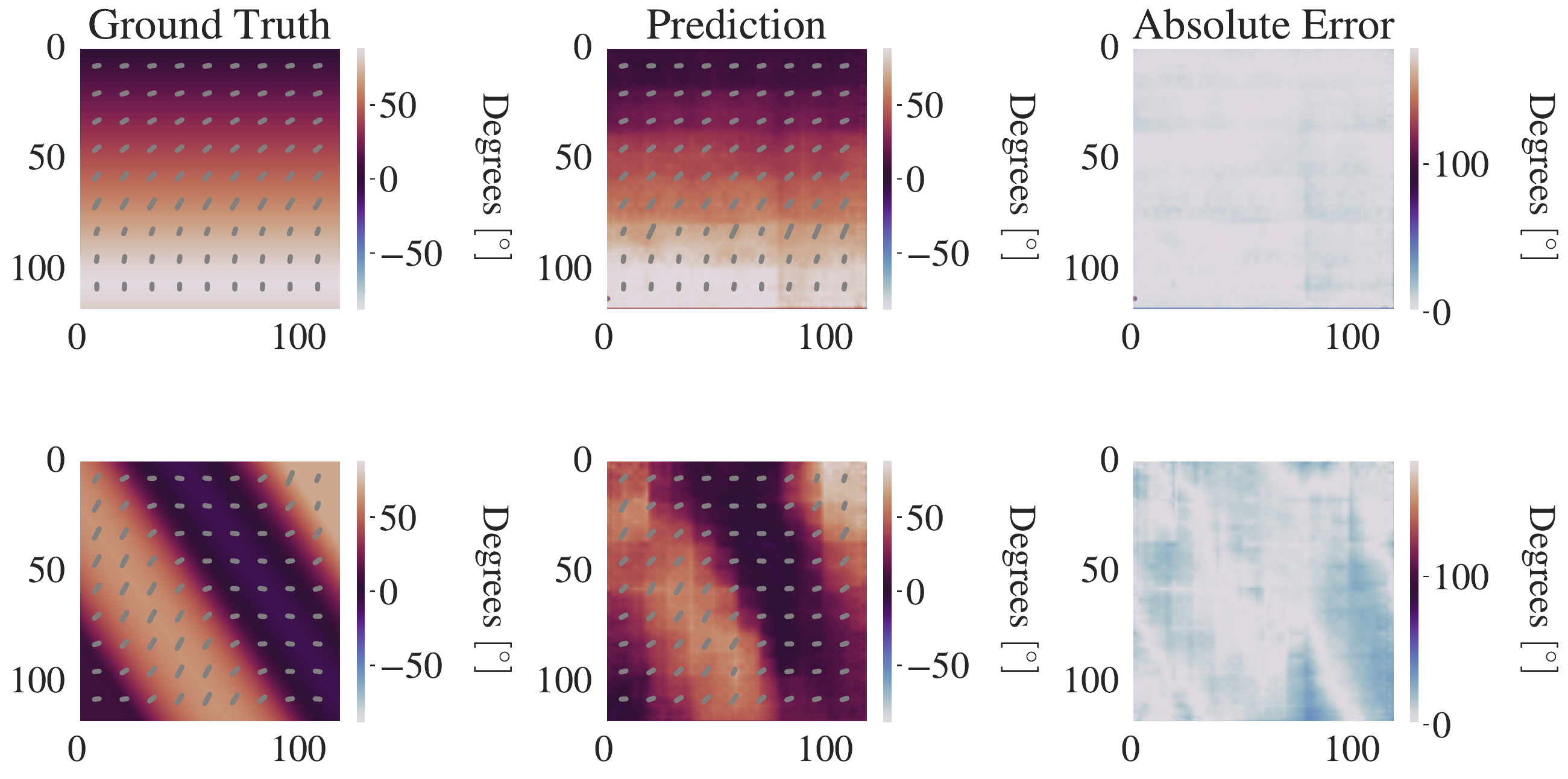}
    \caption[Anisotropic]{Examples of two predicted anisotropic fields (top and bottom). Lines have been superimposed to make the visualization of the prediction easier.}
    \label{fig:7_anisotropic}
\end{figure}

Reconstructed isotropic and anisotropic fields are shown in Figures \ref{fig:7_isotropic} and \ref{fig:7_anisotropic}, respectively. In the heterogeneous isotropic case a Jaccard index of $j = 0.91$ is obtained. To further validate the model, we compare the predicted fields against the average scar map that is produced from our scar map generators, to ensure that the estimator does not simply converge to predict the mean of the \emph{scar maps} set. The predictions are significantly different from the average with $p_{val} = 6 \times 10^{-8}$; as this can be an overestimation of the performance of the network, we present in the next section a more conservative evaluation.

\subsection{Surrogate testing analysis}
\label{subsec:ressurrogates}

For each testing simulation, 100 surrogate fields are produced from the network output. Every surrogate and prediction is compared to the corresponding ground truth conductivity map. The RMSE is used as the discriminating statistic.  We then calculate the percentile of the prediction in the surrogate distribution. The median percentile $P$ for the testing set against the surrogate distributions is $0.0$, and the distribution of $d_{\rm{prediction}}$ compared to $d_{\rm{surrogates}}$ has a significantly smaller mean with a $p_{val} = 0.007$.

\begin{figure}[ht]
    \centering
    \includegraphics[width=0.8\textwidth]{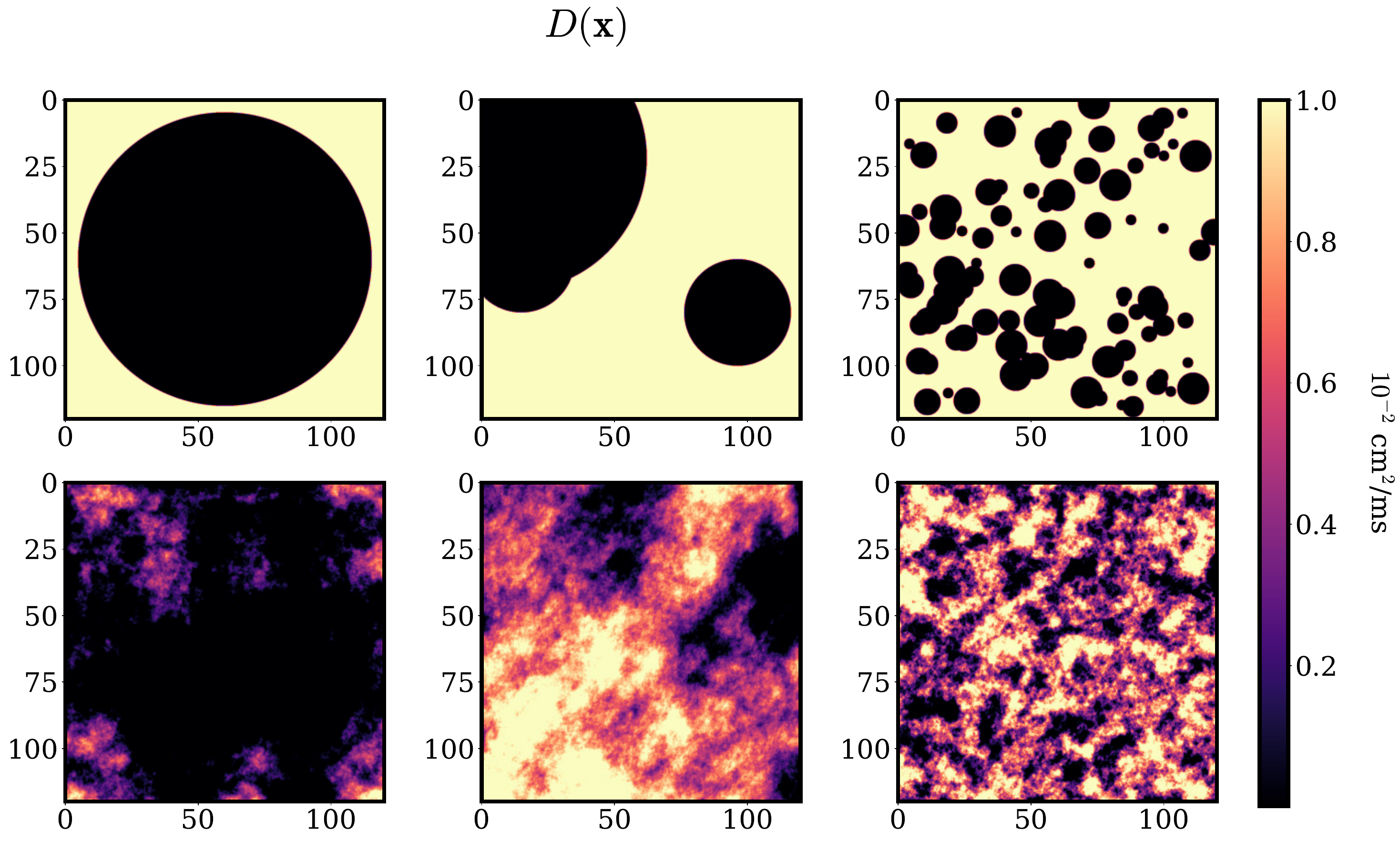}
    \caption[Surrogates]{Examples of autocorrelation-preserving surrogates. \textbf{Top}: Three examples of scar maps, with different degrees of spatial autocorrelation. \textbf{Bottom}: The respective surrogates produced via the DT-CWT method.}
    \label{fig:8_surrogates}
\end{figure}

%% file: 5_discussion.tex
\section{Discussion}
\label{sec:discussion}
 
The results presented in this paper serve a dual purpose. From a machine learning perspective, it provides an initial proof of concept that deep neural networks can be used to solve inverse problems for unsteady, previously unseen, cardiac action potential systems modelled by partial differential equations, such as those contained in the specific cardiac model used in this work.
From a biological perspective, it confirms the hypothesis that electrogram signals can be used to recover structural information about the underlying cardiac substrate in an \textit{in silico} setting despite the level of spatial information content being lower than that of all individual contributing APs within the whole tissue and even while taking into consideration additional biological noise.
This includes identifying compact non-conducting fibrotic regions, and fibre orientation, using a spatial EGM resolution that is consistent with clinical high-density mapping catheters, opening potential avenues for more practical applications.

Furthermore, our method of generating data is on its own useful for further experimentation with EGMs and their relation to tissue structure. As shown in Figure \ref{fig:biological_egms_vs_sim}, we are able to produce synthetic EGMs that resemble biological ones, through the process of enriching them with noise extracted from ex-vivo acquired signals. The solver that we present can be easily adapted to more complex grids and geometries, and the scar maps can be adjusted model different kinds of scar and fibre distributions.

Although similar networks are being increasingly used with impressive performance as solvers, to predict future states of dynamical systems from past states \cite{fotiadis2020}, their ability to solve inverse problems \cite{Wu2020} has not been explored as much, especially for time-dependent PDEs. Similar inverse problems have been solved using a denser electrogram mapping grid and traditional optimization techniques \cite{Abdi2019}, however, the investigated cases have been significantly simpler and required a large number of electrograms. Other methods make use of activation times or APs rather than electrogram recordings \cite{Grandits2021}, \cite{ep_pins} which is the clinical raw data recorded and allows hypothesis-free interrogation by the network. Furthermore, this work allows for the estimation of conductivity fields in unknown domains by relying only on the electrogram signals, in contrast to the other existing methods which interpolate within the same domain, and may require information like system-specific properties \cite{fibre_net}.

The choice of working with electrogram signals instead of the AP is a practical one and for the purposes of future translation; in a biological or clinical setting, contact electrograms can be acquired in contact with the myocardium using mapping catheters. Conversely, directly acquiring the AP propagating through the cardiac muscle requires the use of techniques which, in some cases, are impossible in a clinical setting, such as \textit{ex vivo} optical mapping. As mentioned in section \ref{subsec:ccm}, the density of the electrogram grid is selected, so that each smaller grid of 4 by 4 electrode probes is a representation of a modern state-of-the-art catheter mapping system (HD grid, Abbott Medical). While generating the simulations, a grid discretization of $dx = 0.01 \un{cm}$ is used, while in the EGM grid, the spacing is $dx_{EGM} = 0.4 \un{cm}$, and the output has a discretization of $dx = 0.125 \un{cm}$. Therefore, by considering only the EGM signals, we are working with a coarse grain approximation of the field, with the input grid containing $29^2/96^2=0.91$ or $9.1 \%$ of the spatial information of the output tensor. Given this reduction in information and resolution, the accuracy in the identification of the compact scar regions demonstrates the effectiveness of our approach.

In this \textit{in silico} investigation, the generated data is produced using a specific set of parameters for the Fenton-Karma AP model. We do not consider the performance of the model in predicting conductivity fields from simulations generated from a range of AP parameter values, as this is beyond the scope of this paper. Subsequent experimentation could be conducted to investigate how much impact the range of the model parameters has on the predictive ability of the model. Furthermore, it can be seen quantitatively and qualitatively in section \ref{subsec:res_performance} that isotropic fields with compact patches are better estimated than the fibre orientations of the anisotropic fields. One explanation for this is the observation that the compact fibrotic areas have a greater impact in wave propagation than gradually changing fibre orientations. At this point, it is also not beneficial to increase the complexity of the in-silico simulations. We demonstrate successfully that EGMs can be used by machine learning algorithms to recreate fibre architectures and accurately identify scar patterns. As the next step, implementation on biologically informed fibre orientation fields will verify that the model can perform as well as it does in the case of compact substrate.

Treatment for AF has been so far largely empirical-based, and as we mention, the success rates for ablation are low. Experience and trust on standard, mannually determined procedures is insufficient. If we do not bridge the gap between determinant structure and consequent EP activity, the ability to correct EP activity by further altering the structure is limited. There is not a complete understanding of how beneficial contact EGMs can be in the identification of target regions for ablation, or how much of an improvement they provide over standard procedures such as pulmonary vein isolation (PVI). For a while, many clinicians considered complex fractionated atrial electrograms (CFAEs) to indicate candidate target sites for AF \cite{star_af_original}, although that is now met with skepticism \cite{cfaes,cfaes_2}. There have been studies that suggest that EGM-guided ablation provide no added benefit over PVI \cite{star_af}. In this work, the hypothesis is not that the morphology of EGMs directly indicates candidate regions for ablation, but rather that an array of concurrent EGMs can be used to determine properties of the underlying substrate and therefore indirectly steer ablation targeting of pathophysiological myocardium\cite{cantwell_review} that disrupts the correct propagation of the cardiac electrical signal.

In our work, only unipolar EGMs are considered. The reasoning behind this choice is that the raw signal is used as the input into the deep learning models. Most state-of-the-art techniques that seek to estimate the direction of the wavefront, which are used in the clinic usually consider the bipolar electrogram or the relatively new omnipolar mapping \cite{abbott_1}. The bipolar EGM is the result of the difference of two unipolar EGMs where the electrodes are placed in close proximity to one another, with the usual inter-electrode distance currently being close to $2\un{mm}$. The popularity of the bipolar EGM is based on its filtering properties; when properly positioned and oriented, a bipolar EGM will cancel out far-field signals or other sources of noise \cite{de_bakker} due to simultaneous detection at both electrodes. While widely used for catheter mapping and ablation, the morphology and properties of the bipolar EGM, like its unipolar components, is still not yet entirely understood. It has been suggested that bipolar and unipolar EGMs can be used together for optimal identification of ablation targets \cite{de_bakker}. Should the approach proposed here work in a clinical setting, signals from electrogram arrays will be used for the estimation of scar location. In our model the network considers the juxtaposition of neighbouring electrodes, thereby, while not directly considering bipolar EGMs, may benefit from similar advantages of proximity.

%% file: 6_conclusion.tex
\section{Conclusion}
\label{sec:conclusion}

This work verifies the hypothesis that, in an \textit{in silico} model, electrogram recordings can be used in conjunction with deep neural networks to estimate the conduction properties of the underlying myocardium. Although the electrogram recordings, conductivity and fibre orientation of biological samples are significantly more complex and noisy, the principle of the deep neural network as an inverse solver remains the same. Applying the predictive model to biological and/or clinical settings requires further investigation and data and is beyond the scope of this paper.

\section{Acknowledgements}
\label{sec:acknowledgements}

This work was supported by the Wellcome Trust under Grant 222845/Z/21/Z. For the purpose of open access, the author has applied a CC BY public copyright licence to any Author Accepted Manuscript version arising from this submission.

\section{Author Contributions}
\label{sec:contributions}

K.N., C.C, R.C, and A.B conceptualized the study. K.N. generated the simulated data, built and trained the model and analyzed the results. K.N., and E.P. wrote the software to generate the simulated data. N.P. and R.C. contributed biological expertise. K.N., E.P., C.C., R.C. and A.B. wrote the manuscript.

\section{Declaration of interests}
\label{sec:coi}
The authors declare no competing interests.

\section{Data availability}
\label{sec:data}
Code to reproduce all the simulations, models, and analysis is provided upon request, and will be made available upon acceptance of this manuscript. The dataset of biological EGMs from ex-vivo pig heart experiments can be provided upon request.